\documentclass[runningheads]{llncs}
\usepackage[T1]{fontenc}
\usepackage{graphicx}
\usepackage{gensymb}
\usepackage{amsmath}
\usepackage{longtable}
\usepackage{adjustbox}
\usepackage{array}
\usepackage{booktabs}
\usepackage{pgfplots}
\pgfplotsset{compat=1.18}
\usepackage{pgfplotstable}
\usepackage{tikz}
\newcolumntype{C}[1]{>{\centering\arraybackslash}p{#1}}

\begin{document}
\title{LatentPrintFormer: A Hybrid CNN-Transformer with Spatial Attention for Latent Fingerprint identification}
\titlerunning{LatentPrintFormer for Latent Fingerprint
identification}
\author{Arnab Maity\inst{1,2}\orcidID{0009-0007-7832-4565} \and Manasa \inst{2}\orcidID{0009-0008-3382-5259} \and Pavan Kumar C\inst{2}\orcidID{0000-0002-4907-5412} \and 
Raghavendra Ramachandra\inst{3}\orcidID{0000-0003-0484-3956}
}

\authorrunning{Arnab et al.}
\institute{Ramakrishna Mission Vidyamandira, Belur Math, India \\ \email{arnabmaity.cs@gmail.com}\and Indian Institute of Information Technology (IIIT), Dharwad, India.\\ \email{\{manasa.24phdcs14,pavan\}@iiitdwd.ac.in}    \and
 Norwegian University of Science and Technology (NTNU), Norway.
\\
\email{raghavendra.ramachandra@ntnu.no}\\
%ABC Institute, Rupert-Karls-University Heidelberg, Heidelberg, Germany\\ , pkc@aid.svnit.ac.in, kpu@eced.svnit.ac.in
}
\maketitle              % typeset the header of the contribution
\begin{abstract}
Latent fingerprint identification remains a challenging task due to low image quality, background noise, and partial impressions. In this work, we propose a novel identification approach called LatentPrintFormer. The proposed model integrates a CNN backbone (EfficientNet-B0) and a Transformer backbone (Swin Tiny) to extract both local and global features from latent fingerprints. A spatial attention module is employed to emphasize high-quality ridge regions while suppressing background noise. The extracted features are fused and projected into a unified 512-dimensional embedding, and matching is performed using cosine similarity in a closed-set identification setting. Extensive experiments on two publicly available datasets demonstrate that LatentPrintFormer consistently outperforms three state-of-the-art latent fingerprint recognition techniques, achieving higher identification rates across  Rank-10. 
%evaluate multiple state-of-the-art fingerprint matching algorithms and identify key performance limitations. To address these, we propose a lightweight hybrid architecture that integrates convolutional and transformer-based features with spatial attention to improve the robustness of the representation. The model is trained and tested on a publicly available latent fingerprint dataset, where it demonstrates improved rank-1 to 10 identification accuracy compared to existing methods. On the more challenging LFIW "in-the-wild" dataset, it achieves competitive performance and stands out at higher ranks. This confirms its ability to effectively reduce candidate lists in demanding real-world scenarios.
\keywords{Latent fingerprints  \and Fingerprint identification \and Biometrics} 
\end{abstract}
\section{Introduction}
Fingerprint recognition is widely regarded as one of the most reliable biometric methods, with applications in personal authentication, national ID systems, and forensic investigations. While rolled and slap fingerprints are captured under controlled conditions with high-quality ridge detail, forensic scenarios often involve latent fingerprints unintentionally left on surfaces and later developed using chemical or physical methods.

Latent fingerprints are typically partial, noisy, and low in contrast, often affected by background clutter, occlusions, and distortions. These factors make matching latent prints to rolled ones significantly more difficult. Traditionally, latent fingerprint recognition has relied on minutiae-based algorithms that extract and compare local ridge features such as bifurcations and ridge endings using geometric matching.

In recent years, deep learning has gained significant traction in addressing the challenges of latent fingerprint matching. Grosz and Jain demonstrated the benefits of combining local and global features through the fusion of embeddings~\cite{grosz2023latent}, while their AFR-Net introduced attention mechanisms to enhance key ridge features in noisy prints~\cite{grosz2023afr}. Other efforts focus on image enhancement; for example, Yadav et al. employed fluorescent carbon quantum dots combined with machine learning to improve latent print visualization and matching~\cite{yadav2024harnessing}. These studies reflect ongoing progress toward more robust and automated latent fingerprint recognition systems.

Despite recent advances, latent fingerprint matching remains highly challenging due to low image quality, noise, and partial impressions. In this study, we evaluate several state-of-the-art fingerprint matching algorithms to identify their limitations on latent prints. To address these challenges, we propose a lightweight hybrid architecture that combines convolutional and transformer-based features, enhanced by a spatial attention mechanism for more reliable representation.

\section{Related Work}
Recent advancements in latent fingerprint recognition span several key areas of the biometric pipeline, including segmentation, minutiae extraction, quality assessment, and matching. Table~\ref{tab:related_works} summarizes notable methods and their distinctive features, covering both traditional and deep learning-based approaches. For non-overlapped latent fingerprint segmentation, Cao et al. ~\cite{cao2014segmentation} introduced a coarse-to-fine dictionary learning framework that leverages TV decomposition, ridge structure modeling, and SSIM-based scoring, leading to a multi-stage Gabor based enhancement pipeline. Sankaran et al. ~\cite{sankaran2017adaptive}  proposed an Adaptive Random Decision Forest (RDF) architecture with saliency-guided segmentation and RELIEF-based feature selection to improve ridge extraction. Deep learning models have also shown promise: Stojanović et al. ~\cite{stojanovic2016fingerprint} used CNN-based ROI classification with neighborhood smoothing, while Nguyen et al. ~\cite{nguyen2018automatic} developed SegFinNet, a fully convolutional network combining visual attention with fusion voting for one-shot, non-patch-based segmentation. Additionally, Zhang et al. \cite{zhang2012latent}presented a Directional TV (DTV) method using an augmented Lagrangian solver for orientation-aware decomposition and structured noise removal~.

In minutiae extraction, Su and Srihari  ~\cite{su2010latent} applied Gaussian Process Regression on orientation maps to estimate core points with being affected by rotation and translation. Sankaran et al. ~\cite{sankaran2014latent}  created a stacked denoising sparse autoencoder (SDSAE) followed by binary classifiers to differentiate minutiae from non-minutiae areas, using ten print patches for training. Tang et al. ~\cite{tang2017latent} suggested a hybrid architecture combining FCN and CNN that uses multi-task learning for minutiae localization and orientation estimation without need for preprocessing. Also Tang et al. ~\cite{tang2017fingernet} proposed FingerNet that takes a comprehensive approach by integrating enhancement, orientation estimation, and minutiae detection into a single end-to-end CNN-based pipeline.

Numerous methods have been developed for latent fingerprint comparison/identification. Deshpande et al. ~\cite{deshpande2020cnnai} proposed CNNAI, a CNN-based matcher that uses local minutiae descriptors and hash indexing for efficient search. Paulino et al. ~\cite{paulino2012latent} introduced a non-deep-learning approach combining Minutia Cylinder Codes (MCC) with a Hough transform voting mechanism for alignment-free matching. Cao and Jain ~\cite{cao2018automated} presented a ConvNet-based template matcher that leverages both minutiae and texture features, demonstrating robustness to incomplete impressions. Grosz et al. ~\cite{grosz2022minutiae} developed a minutiae-guided Vision Transformer (ViT) that uses minutiae heatmaps and grayscale ridge images to produce fixed-length embeddings, enabling over 2.5 million comparisons per second. Jain and Feng ~\cite{jain2010latent} introduced the Extended Feature Fusion Matcher, which incorporates manually annotated extended features based on the CDEFFS standard. Pan et al. ~\cite{pan2024latent}proposed the Dense Minutia Descriptor (DMD), a dual-branch CNN that generates spatially aware 3D descriptors with score normalization across overlapping regions. Finally, Cao et al. ~\cite{cao2019end} presented an end-to-end latent search pipeline combining autoencoders, CNNs, and template fusion for fast retrieval using product quantization.

\begin{longtable}{|C{0.22\textwidth}|C{0.25\textwidth}|C{0.48\textwidth}|}
\caption{Overview of existing latent fingerprints techniques}
\label{tab:related_works} \\
\toprule
\textbf{Method} & \textbf{Architecture} & \textbf{Key Features} \\
\midrule

\multicolumn{3}{|c|}{\textit{\textbf{Non Overlapped Latent Fingerprint Segmentation}}} \\
\midrule
Coarse-to-fine Dictionary~\cite{cao2014segmentation} & Dictionary Learning + Sparse Coding & TV decomposition, ridge structure dictionary, SSIM-based quality scoring, coarse-to-fine enhancement with Gabor filters \\
\midrule
Adaptive RDF~\cite{sankaran2017adaptive} & Random Decision Forest (RDF) & Saliency-guided segmentation, RELIEF-based feature selection \\
\midrule
Deep Learning Based ROI Segmentation~\cite{stojanovic2016fingerprint} & CNN & Block-wise ROI classification, Neighborhood Smoothing \\
\midrule
SegFinNet~\cite{nguyen2018automatic} & FCNN & Visual attention, fusion voting, non-patch based, one-shot segmentation \\
\midrule
DTV~\cite{zhang2012latent} & Directional TV + Augmented Lagrangian Solver & Orientation-aware decomposition, structure noise removal \\
\midrule

\multicolumn{3}{|c|}{\textit{\textbf{Overlapped Latent Fingerprint Segmentation}}} \\
\midrule
Orientation-Guided Separation~\cite{chen2011separating} & Fourier Transform + Relaxation Labeling & Singularity-guided enhancement, relaxation labeling \\
\midrule
Model-Based Overlap Separation~\cite{zhao2012model} & Orientation Field Modeling + Polynomial Regression & Iterative orientation field reconstruction, model-based prediction and regularization \\
\midrule
ANFIS~\cite{jeyanthi2016efficient} & ANFIS Classifier + Feature Engineering + Orientation Field Reconstruction & Fuzzy inference, orientation-guided separation, feature-based classification \\
\midrule
Neural Network-Based Separation~\cite{stojanovic2017novel} & Neural Network and Fourier Analysis & Local orientation modeling, iterative refinement and smoothing \\
\midrule

\multicolumn{3}{|c|}{\textit{\textbf{Minutiae Extraction}}} \\
\midrule
Core Point Prediction via Gaussian Processes~\cite{su2010latent} & Gaussian Process Regression on Orientation Maps & Probabilistic core point estimation, rotation and translation invariant \\
\midrule
Minutiae Detection via SDSAE~\cite{sankaran2014latent} & Stacked Denoising Sparse Autoencoder (SDSAE) + Binary Classifiers & Learns minutiae/non-minutiae descriptors from tenprint patches, patch-wise binary classification of latent prints \\
\midrule
Minutiae Extraction via FCN + CNN~\cite{tang2017latent} & FCN + CNN & Multi-task loss for location and orientation, end-to-end minutiae detection without preprocessing \\
\midrule
FingerNet~\cite{tang2017fingernet} & CNN & End-to-end enhancement, orientation estimation \\
\midrule

\multicolumn{3}{|c|}{\textit{\textbf{Quality Assessment}}} \\
\midrule
LFIQ~\cite{yoon2013lfiq} & Feature-based scoring system using ridge quality & Ridge quality estimation via ridge amplitude and continuity, finger position estimation using orientation field and curvature-based core detection \\
\midrule
Deep Latent Quality Estimator~\cite{ezeobiejesi2018latent} & Deep Neural Network & Patch-wise quality classification, automated segmentation using 8x8 patch classification \\
\midrule

\multicolumn{3}{|c|}{\textit{\textbf{Verification/Identification}}} \\
\midrule
CNNAI~\cite{deshpande2020cnnai} & CNN & Local minutiae-based, rotation-scale invariant, hash indexing \\
\midrule
MCC + Hough Transform~\cite{paulino2012latent} & Non-DL & Minutiae cylinder codes, alignment-based \\
\midrule
ConvNets~\cite{cao2018automated} & CNN & Minutiae descriptor-based matching with learned features \\
\midrule
Minutiae-Guided ViT~\cite{grosz2022minutiae} & Vision Transformer & Minutiae-guided attention maps for fixed-length embedding generation \\
\midrule
Extended Feature Fusion Matcher~\cite{jain2010latent} & Non-DL & Uses manually marked extended features, leverages CDEFFS-standard cues \\
\midrule
Dense Minutia Descriptor (DMD)~\cite{pan2024latent} & Dual-branch CNN & Dense 3D descriptors with spatial structure, matching via overlapping region and score normalization \\
\midrule
End-to-End Latent Search~\cite{cao2019end} & Autoencoder + CNN + Template Fusion & Combines minutiae and texture templates, fast search via product quantization \\
\midrule
\textbf{Proposed Method: LatentPrintFormer} & \textbf{CNN + Swin Transformer} & \textbf{EfficientNet-B0 + Swin Tiny, Spatial attention, embedding-based similarity using cosine metric} \\
\bottomrule
\end{longtable}

Our review of state-of-the-art latent fingerprint recognition reveals a clear divide in advanced matching systems. CNN-based models like FingerNet~\cite{tang2017fingernet} excel at capturing local textures and minutiae-like features, while Transformer-based models such as the Minutiae-Guided ViT~\cite{grosz2022minutiae} effectively model global context and long-range dependencies. However, few approaches successfully integrate both. Existing methods typically rely on a single architecture or combine features from separate, often complex modules. This highlights the need for a lightweight, end-to-end hybrid architecture that unifies CNNs and Transformers to leverage both local and global representations. Moreover, although attention mechanisms have been explored~\cite{grosz2023afr}, few models incorporate spatial attention to suppress background noise and emphasize ridge-rich regions. Our work addresses this architectural and functional gap with a model that is both efficient and robust. The following are the main contributions of this work: 
\begin{itemize}
    \item We propose \textbf{LatentPrintFormer}, a novel hybrid architecture that combines a CNN backbone (EfficientNet-B0) with a Transformer backbone (Swin Tiny) for latent fingerprint identification.
    
    \item A \textbf{Spatial Attention Module} is integrated to enhance discriminative regions and suppress background noise, improving robustness in noisy and partial latent prints.
    
    \item The proposed architecture produces a unified 512-dimensional embedding optimized with ArcFace loss and cosine similarity, enabling accurate closed-set identification.
    
    \item We conduct extensive experiments on two publicly available datasets (IIITD and LFIW) and benchmark our method against three state-of-the-art latent fingerprint recognition systems (NBIS \cite{nistnbis}, MCC \cite{cappelli2010minutia}, MSU-LatentAFIS~\cite{cao2019end}).
    
    \item Results demonstrate that \textbf{LatentPrintFormer} consistently achieves higher rank-10 identification accuracy, particularly in challenging forensic conditions.
\end{itemize}

The rest of the paper is organised as follows: Section \ref{sec:Prop} discuss the proposed LatentPrintFormer for latent fingerprint identification, Section \ref{sec_Exp} discuss the datasets and experimental results and finally Section \ref{sec:conc} draws the conclusion. 
%%%%%%%%%%%%%%%%%%%%%%%%%%%%%%%%%%%%%%
\section{Proposed LatentPrintFormer for latent fingerprint identification}
\label{sec:Prop}
\begin{figure}
    \centering
      \resizebox{\columnwidth}{!}{%
    \includegraphics[angle=90]{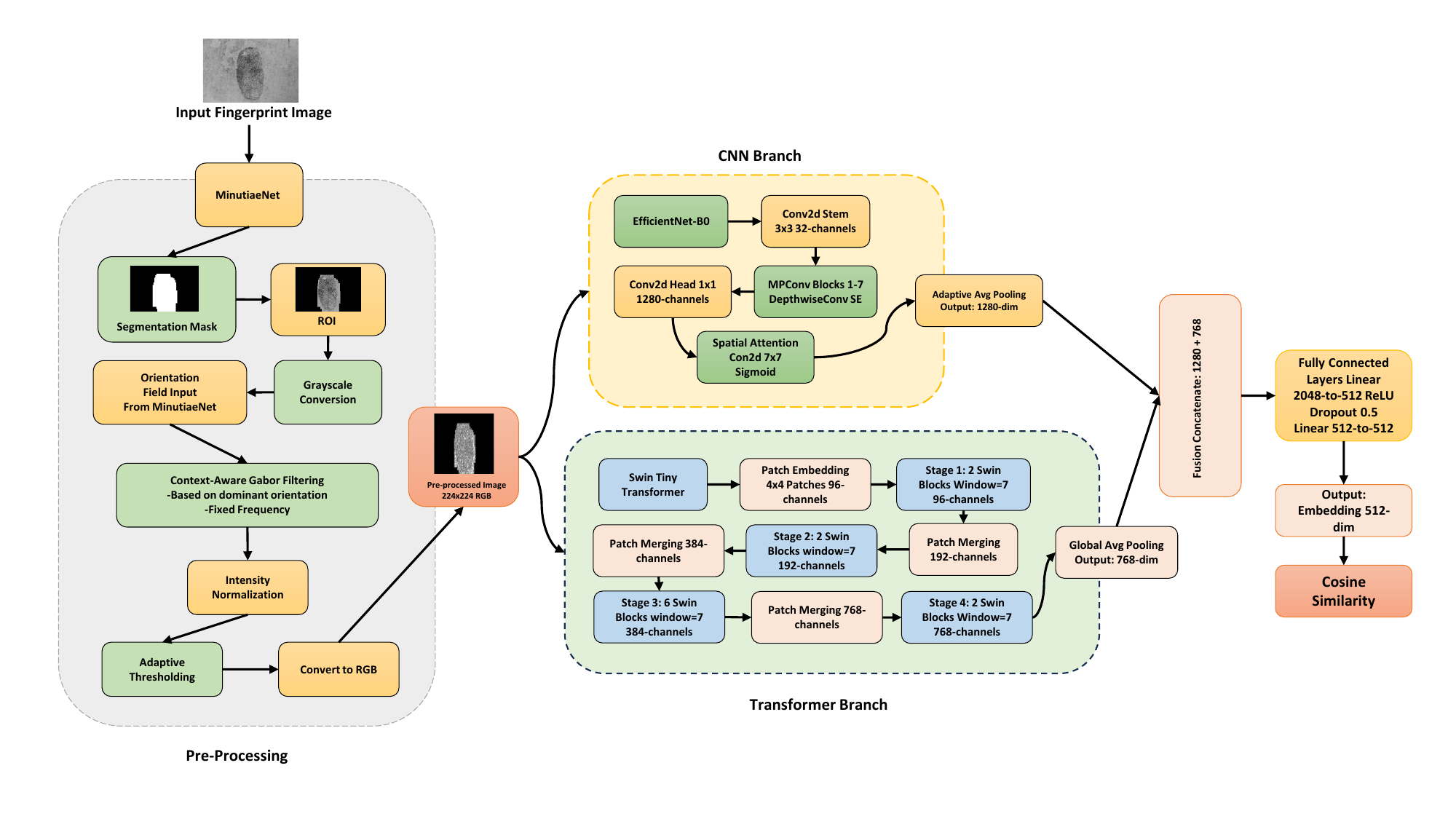}}
    \caption{Block diagram of the proposed LatentPrintFormer Architecture}
    \label{fig:architecture}
\end{figure}
This work presents a hybrid deep learning architecture, called LatentPrintFormer, which combines Convolutional Neural Networks (CNNs) and Transformer-based models for reliable latent fingerprint identification. The proposed method addresses key challenges in uncontrolled environments, such as partial impressions, noise, and poor-quality latent prints. LatentPrintFormer leverages representation learning, spatial attention, and embedding-based similarity computation to enhance identification accuracy.

\subsection{Design Rationale}
Reliable latent fingerprint identification requires extracting features that are consistent for the same identity and distinct across different individuals. Traditional methods relying on handcrafted features—such as minutiae, ridge flow, and texture often fail under conditions of noise, occlusion, and distortion. In contrast, deep learning models offer a data-driven alternative. Convolutional Neural Networks (CNNs) effectively capture local patterns like ridge endings and bifurcations, while Transformers excel at modeling global context through self-attention, making them well-suited for interpreting partial or degraded fingerprint images.

To leverage the strengths of both architectures, our system combines a CNN backbone (EfficientNet) with a Transformer backbone (Swin Transformer). This hybrid design is further enhanced by a spatial attention mechanism that highlights the most informative regions of the latent fingerprint.
\subsection{LatentPrintFormer Network Architecture}
The proposed LatentPrintFormer architecture includes four main components: a segmentation and pre-processing, a CNN feature extractor, a Transformer-based global context encoder, and a feature fusion module that combines the gathered information into a single embedding space.

\subsection{Segmentation and Preprocessing}
Latent fingerprint images often contain noisy, low-contrast ridge patterns and complex, cluttered backgrounds, making it difficult to extract meaningful features without first isolating the region of interest. To address this, we use the MinutiaeNet segmentation module, a deep learning framework designed for fingerprint segmentation and minutiae detection~\cite{Nguyen_MinutiaeNet}.
In practice, segmentation may result in multiple disconnected foreground regions, especially in partial or degraded prints. To ensure consistency and suppress noise, we perform connected component analysis on the binary mask and retain only the largest connected area, assuming that it corresponds to the actual fingerprint region.

This final mask is applied to the original latent image to eliminate irrelevant regions and focus on ridge-bearing areas. This segmentation step enhances the quality and reliability of feature extraction by ensuring that only valid fingerprint data is processed. The segmented region of interest (ROI) is then converted to grayscale, and context-aware Gabor filtering, guided by the dominant local orientation from MinutiaeNet, is applied using a fixed frequency to enhance ridge clarity.

Subsequently, the image is normalized to maintain consistent contrast and brightness, followed by adaptive thresholding to emphasize ridge-valley structures and suppress residual background noise. Finally, the processed grayscale image is converted into a 3 channel RGB format and resized to 224×224 pixels to align with the input requirements of the hybrid CNN-Transformer architecture. This comprehensive preprocessing pipeline ensures that both local and global features can be extracted more effectively by improving ridge visibility and minimizing noise.  
\subsubsection{CNN Branch - Local Feature Extraction} 
The first part of LatentPrintFormer uses a pretrained EfficientNet-B0 \cite{tan2019efficientnet} as the CNN backbone. EfficientNet is a family of convolutional architectures optimized for both accuracy and computational efficiency by uniformly scaling depth, width, and resolution. We chose EfficientNet-B0 \cite{tan2019efficientnet} over other CNNs due to its lightweight design and strong performance on visual recognition tasks with fewer parameters. In our implementation, the classification layers are removed, and the convolutional trunk is retained to extract a high-resolution spatial feature map with 1280 channels. This representation effectively captures fine-grained ridge patterns crucial for fingerprint discrimination.

\subsubsection{Spatial Attention Mechanism}
Given the varying quality of latent fingerprint impressions, especially in forensic scenarios where not all regions of the image are equally informative. While EfficientNet effectively captures local features, it treats all spatial regions uniformly, which may dilute the influence of critical ridge structures. To address this, we incorporate a Spatial Attention Module that selectively emphasizes the most relevant areas. Specifically, a learnable $7 \times 7$ convolution followed by a sigmoid activation generates an attention map that highlights high-quality ridge regions while suppressing background noise and irrelevant details. This refined focus enhances the local feature representation, which is then aggregated using global average pooling to produce a compact, fixed-length descriptor.

\subsubsection{Transformer Branch - Global Context Modeling}
To complement the local feature extraction of EfficientNet \cite{tan2019efficientnet}, we use a Swin Transformer as the second feature extraction pathway to capture global context and long-range dependencies within fingerprint images. This is particularly important in latent fingerprints, where ridge patterns may be fragmented or distorted. The Swin Transformer processes the image by dividing it into non-overlapping patches and applying hierarchical self-attention within shifted windows. This mechanism enables the model to efficiently capture spatial relationships across distant regions while maintaining computational efficiency. The output is a 768-dimensional embedding that encodes the global structure of the fingerprint.

\subsubsection{Feature Fusion and Embedding Projection}
The outputs from the CNN and Transformer branches, with dimensions of 1280 and 768 respectively, are concatenated and passed through a fully connected projection head. This head consists of two linear layers with ReLU activation and dropout regularization, which help improve generalization. The projection reduces the combined feature vector to a compact 512-dimensional embedding. This embedding serves as the final fingerprint representation used for identification and matching.

\subsection{Comparison}
During testing, each fingerprint image is transformed into a 512-dimensional embedding vector. These embeddings are then compared using cosine similarity, where a higher score indicates a greater likelihood that the fingerprints belong to the same identity. This approach supports rank-based fingerprint identification and enables effective candidate retrieval.
\subsection{Additional Implementation Details}
The LatentPrintFormer model is trained from start to finish using the ArcFace loss, with a margin of 0.5 and a scale of 64. The optimizer is Adam, featuring a learning rate of 1e-4, a weight decay of 1e-5, and a batch size of 16. Input images are resized to 224x224 and normalized using a mean and standard deviation of [0.5, 0.5, 0.5]. During training, data augmentation techniques include random rotation of $\pm$15\degree, horizontal flipping with a probability of 0.5, brightness and contrast adjustment of $\pm$0.1, Gaussian blur with kernel size of 3. The concatenated features go through a two-layer projection head, which includes linear transformations with ReLU activation and a dropout rate of 0.5 to create the final embedding.
%%%%%%%%%%%%%%%%%%%%%%%%%%%%%%%%%%%%%%%%%%%%%%%%%%%%%%%%%
\section{Experiments and Results}
\label{sec_Exp}
This section presents the experimental setup and results for the proposed and baseline methods in latent fingerprint identification. We begin by describing the datasets used, followed by the evaluation protocol and a discussion of the quantitative results.

\begin{figure}
    \centering
      \resizebox{\columnwidth}{!}{%
    \includegraphics{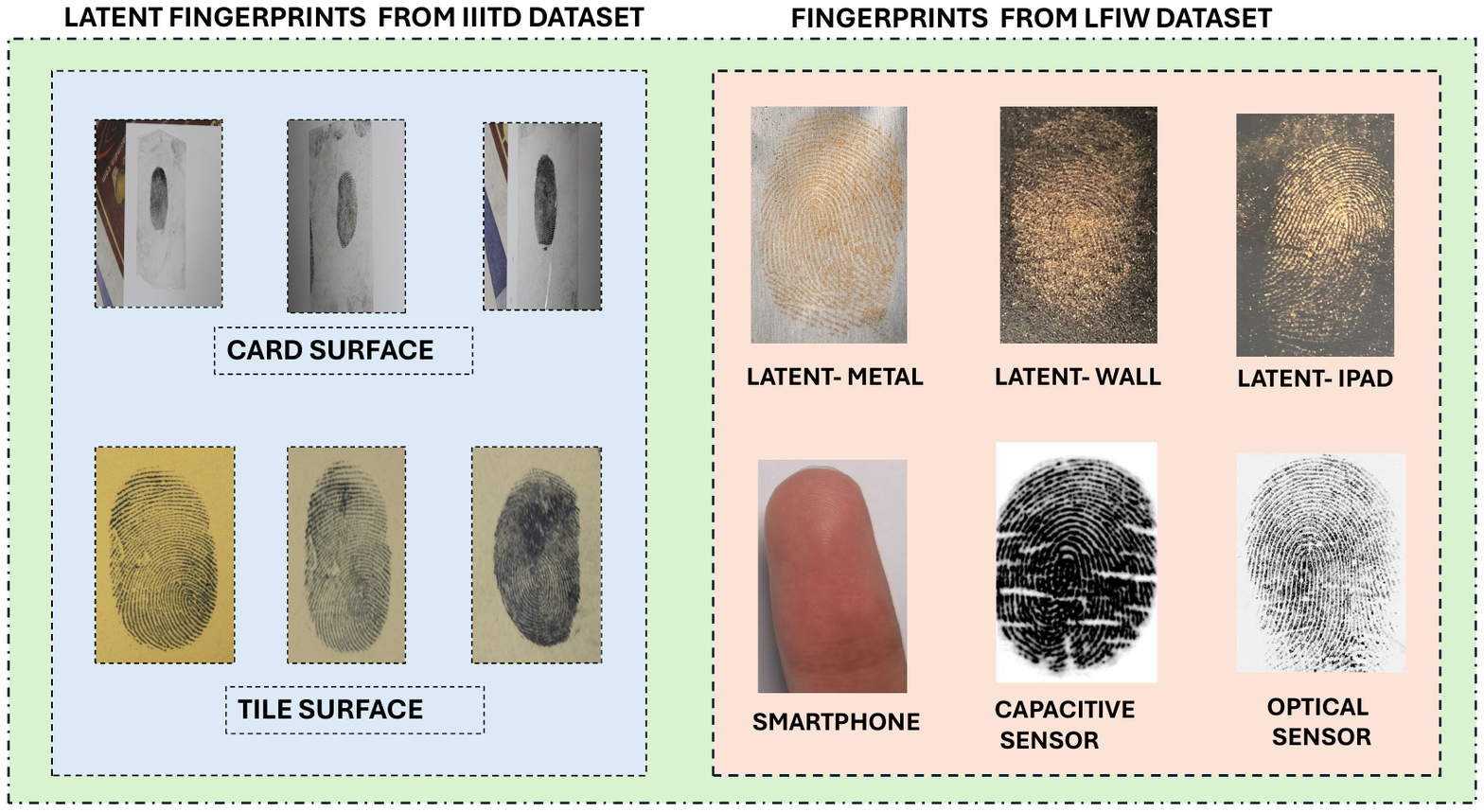}}
    \caption{Example of IIITD and LFIW datasets}
    \label{fig:Dbs}
\end{figure}

\subsection{Dataset and Evaluation Protocol}
In this work, we use two publicly available datasets: the IIITD-Latent Fingerprint Database~\cite{Sankaran2011matching} and the LFIW database~\cite{liu2024latent}. The IIITD dataset contains latent fingerprints from 15 subjects, each contributing 10 fingers. Prints were lifted using black powder and brush on two backgrounds such as card and tile that are captured under semi-controlled conditions with a Canon EOS 500D camera at a resolution of 4752×3168 pixels. Multiple impressions were recorded per background to introduce intra-class variation.

The LFIW dataset includes 6000 images from 60 subjects and is divided into six subsets based on capture method and surface type:
(a) \textbf{R-opt}: Optical sensor reference prints,
(b) \textbf{R-cap}: Capacitive sensor reference prints,
(c) \textbf{Smt}: Fingerphotos from a smartphone,
(d) \textbf{L-wall}: Latents lifted from a wall,
(e) \textbf{L-ipad}: Latents lifted from an iPad, and
(f) \textbf{L-alum}: Latents lifted from aluminum foil.
Figure \ref{fig:Dbs} illustrates the example of images from IIITD-Latent Fingerprint Database~\cite{Sankaran2011matching} and the LFIW database~\cite{liu2024latent}. 

\subsubsection{Evaluation Protocol}
This section outlines the evaluation protocols used to benchmark the performance of the proposed and existing latent fingerprint identification methods.

\textbf{Experiment \#1:} This experiment is designed to evaluate performance on the IIITD-Latent Fingerprint Database using a closed-set identification protocol. A gallery of 150 rolled fingerprints was constructed, and 1046 latent fingerprint instances were used as probes. Each probe was matched against all gallery entries, forming a one-to-many identification scenario.

We use the same set of subjects for both gallery and probe to simulate a closed-set forensic search, where the true identity is assumed to be present in the enrolled database. This setting is particularly relevant for controlled evaluations and ensures a consistent identity space during training and testing, enabling the model to focus on handling latent-specific distortions rather than subject variability.

\textbf{Experiment \#2:} 
In this experiment, we evaluate the generalization capability of the proposed LatentPrintFormer under realistic conditions. The model is trained exclusively on the IIITD dataset and tested on the LFIW dataset. The gallery consists of all R-opt reference prints from 60 subjects, while the probe set includes all other fingerprint types (Smt, L-wall, L-ipad, and L-alum). This protocol poses a challenging cross-sensor and cross-modality matching task, demonstrating the model's robustness beyond the training domain.

\subsection{Result and Discussion}
This section presents and discusses the quantitative results of the proposed and baseline methods across both experimental protocols. We use closed-set identification with Rank-10 accuracy, defined as the percentage of probe samples for which the correct identity appears within the top 10 matches. Higher Rank-10 values indicate better identification performance.

\subsubsection{Results on Experiment \#1}

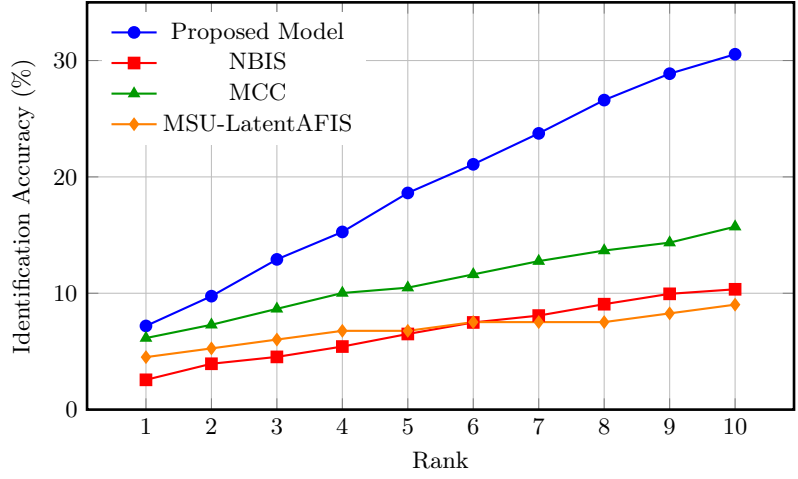
\begin{figure}
\centering
\begin{tikzpicture}
\begin{axis}[
    width=0.9\textwidth,
    height=7cm,
    xlabel={Rank},
    ylabel={Identification Accuracy (\%)},
    xtick={1,2,...,10},
    grid=major,
    legend style={
        at={(0.02,0.98)},
        anchor=north west,
        draw=none,
        fill=white,
        font=\small
    },
    ymin=0, ymax=35,
    line width=1pt,
    mark size=2pt,
    every axis plot/.append style={thick}
]

% Data for each system
\addplot[color=blue, mark=*] coordinates {
    (1,7.19) (2,9.75) (3,12.91) (4,15.27) (5,18.62)
    (6,21.08) (7,23.74) (8,26.60) (9,28.87) (10,30.54)
};
\addlegendentry{Proposed Model}

\addplot[color=red, mark=square*] coordinates {
    (1,2.56) (2,3.94) (3,4.53) (4,5.42) (5,6.50)
    (6,7.49) (7,8.08) (8,9.06) (9,9.95) (10,10.34)
};
\addlegendentry{NBIS}

\addplot[color=green!60!black, mark=triangle*] coordinates {
    (1,6.15) (2,7.29) (3,8.66) (4,10.02) (5,10.48)
    (6,11.62) (7,12.76) (8,13.67) (9,14.35) (10,15.72)
};
\addlegendentry{MCC}

\addplot[color=orange, mark=diamond*] coordinates {
    (1,4.51) (2,5.26) (3,6.02) (4,6.77) (5,6.77)
    (6,7.52) (7,7.52) (8,7.52) (9,8.27) (10,9.02)
};
\addlegendentry{MSU-LatentAFIS}

\end{axis}
\end{tikzpicture}
\caption{Plot of Rank-10 Identification Accuracy on Experiment \#1}
\label{fig:rank-n-plot}
\end{figure}

\begin{table*}
\centering
\caption{Rank-N Identification Accuracy (\%) on the \textbf{Experiment \#1}}
\vspace{0.6em}
\label{tab:rank-n-iiit}
\begin{tabular}{|c|c|c|c|c|}
\hline
\textbf{Rank} & \textbf{Proposed Model} & \textbf{NBIS} \cite{nistnbis} & \textbf{MCC} \cite{cappelli2010minutia} & \textbf{MSU-LatentAFIS} \cite{cao2019end} \\
\hline
1  & \textbf{7.19}  & 2.56 & 6.15 & 4.51 \\
2  & 9.75           & 3.94 & 7.29 & 5.26 \\
3  & 12.91          & 4.53 & 8.66 & 6.02 \\
4  & 15.27          & 5.42 & 10.02 & 6.77 \\
5  & 18.62          & 6.50 & 10.48 & 6.77 \\
6  & 21.08          & 7.49 & 11.62 & 7.52 \\
7  & 23.74          & 8.08 & 12.76 & 7.52 \\
8  & 26.60          & 9.06 & 13.67 & 7.52 \\
9  & 28.87          & 9.95 & 14.35 & 8.27 \\
10 & \textbf{30.54} & 10.34 & 15.72 & 9.02 \\
\hline
\end{tabular}
\end{table*}

Figure~\ref{fig:rank-n-plot} presents a visual comparison of Rank-N identification accuracy for all four systems on the IIITD dataset. Our proposed model, represented by the blue line, consistently outperforms all baselines across ranks 1 to 10. The performance gap is evident at Rank-1 and widens with increasing rank, indicating that the representation of characteristics of our model is more effective in obtaining the correct identity.

Table~\ref{tab:rank-n-iiit} provides the detailed accuracy values. At rank 1, LatentPrintFormer achieves 7. 19\% accuracy and outperforming MCC (6.15\%), MSU-LatentAFIS (4.51\%) and NBIS (2.56\%). At Rank-10, it reaches 30.54\%, nearly double MCC (15.72\%) and more than triple MSU-LatentAFIS (9.02\%).

These results clearly demonstrate the effectiveness of our hybrid CNN-Transformer architecture in improving latent fingerprint identification performance.

\begin{figure}
\centering
\begin{tikzpicture}
\begin{axis}[
    width=0.9\textwidth,
    height=7cm,
    xlabel={Rank},
    ylabel={Identification Accuracy (\%)},
    xtick={1,2,...,10},
    grid=major,
    legend style={
        at={(0.02,0.98)},
        anchor=north west,
        draw=none,
        fill=white,
        font=\small,
        legend columns=1
    },
    ymin=0, ymax=20,
    line width=1pt,
    mark size=2pt,
    every axis plot/.append style={thick}
]

% Data for each system
\addplot[color=blue, mark=*] coordinates {
    (1,2.86) (2,4.28) (3,6.30) (4,8.08) (5,9.73)
    (6,10.91) (7,12.93) (8,14.88) (9,16.73) (10,18.48)
};
\addlegendentry{Proposed Model}

\addplot[color=red, mark=square*] coordinates {
    (1,5.52) (2,6.19) (3,6.49) (4,6.86) (5,7.09)
    (6,7.39) (7,7.69) (8,7.73) (9,7.86) (10,7.93)
};
\addlegendentry{NBIS}

\addplot[color=green!60!black, mark=triangle*] coordinates {
    (1,3.50) (2,4.50) (3,6.00) (4,7.50) (5,8.90)
    (6,10.20) (7,11.80) (8,13.03) (9,15.34) (10,16.56)
};
\addlegendentry{MCC}

\addplot[color=orange, mark=diamond*] coordinates {
    (1,4.80) (2,5.50) (3,6.80) (4,7.20) (5,7.50)
    (6,7.80) (7,8.10) (8,8.40) (9,8.70) (10,9.00)
};
\addlegendentry{MSU-LatentAFIS}

\end{axis}
\end{tikzpicture}
\caption{Plot of Rank-10 Identification Accuracy Comparison on Experiment \#2}
\label{fig:rank-n-plot-lfiw}
\end{figure}
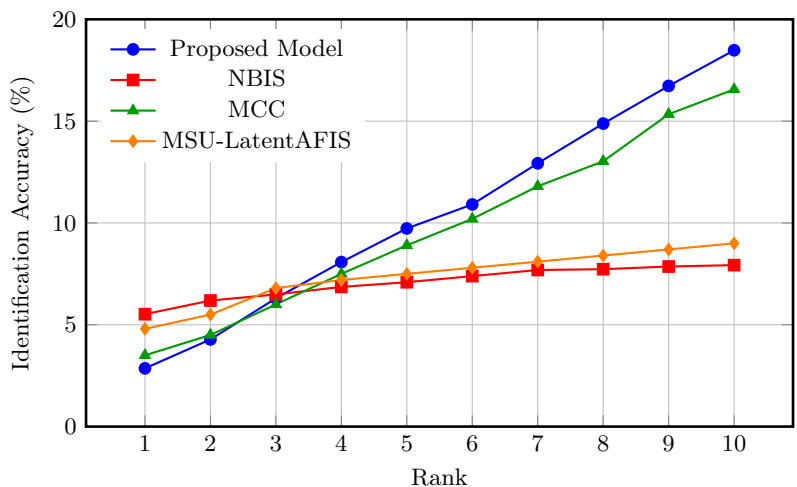

\begin{table*}
\centering
\caption{Rank-N Identification Accuracy (\%)  on \textbf{Experiment \#2}}
\vspace{0.6em}
\label{tab:rank-n-lfiw}

\begin{tabular}{|c|c|c|c|c|}
\hline
\textbf{Rank} & \textbf{Proposed Model} & \textbf{NBIS} \cite{nistnbis} & \textbf{MCC} \cite{cappelli2010minutia} & \textbf{MSU-LatentAFIS}\cite{cao2019end} \\
\hline
1  & 2.86           & 5.52 & 3.50 & 4.80 \\
2  & 4.28           & 6.19 & 4.50 & 5.50 \\
3  & 6.30          & 6.49 & 6.00 & 6.80 \\
4  & 8.08          & 6.86 & 7.50 & 7.20 \\
5  & 9.73         & 7.09 & 8.90 & 7.50 \\
6  & 10.91          & 7.39 & 10.20 & 7.80 \\
7  & 12.93          & 7.69 & 11.80 & 8.10 \\
8  & 14.88          & 7.73 & 13.03 & 8.40 \\
9  & 16.73          & 7.86 & 15.34 & 8.70 \\
10 & \textbf{18.48} & 7.93 & 16.56 & 9.00 \\
\hline
\end{tabular}
\end{table*}

\begin{table}[htbp]
\caption{Quantitative results of the ablation study across LatentPrintFormer model variants}
\label{tab:ablation-study}
\resizebox{\columnwidth}{!}{%
\begin{tabular}{|c|c|c|c|}
\hline
\textbf{CNN} & \textbf{Spatial Attention} & \textbf{Transformer} & \textbf{Rank 10 Identification Accuracy} \\ \hline
Yes & No  & No  & 10.94\% \\ \hline
Yes & Yes & No  & 16.95\% \\ \hline
Yes & Yes & Yes & 30.54\% \\ \hline
\end{tabular}%
}
\end{table}
%Figure~\ref{fig:rank-n-plot} shows a visual comparison of the Rank-N identification accuracy for all four systems on the IIITD Database. The graph illustrates that our proposed model, represented by the blue line, consistently performs better than all baseline systems for every rank from 1 to 10. The performance gap is clear even at Rank-1; more importantly, it grows as the rank increases. This trend strongly indicates that our model's feature representation is more effective at distinguishing identities, allowing it to better retrieve the correct identity.

%For a detailed quantitative analysis, Table~\ref{tab:rank-n-iiit} provides the exact accuracy percentages obtained on the IIITD Database. The numbers confirm our method's superiority. At Rank-1, our model achieves an identification accuracy of 7.19\%. This is a significant improvement over the next-best system, MCC, which has an accuracy of 6.15\%, and it is substantially better than NBIS at 2.56\% and MSU-LatentAFIS at 4.51\%. The performance advantage becomes even clearer at higher ranks. For example, at Rank-10, our model reaches an accuracy of 30.54\%. This is nearly double the performance of MCC at 15.72\% and more than triple that of MSU-LatentAFIS at 9.02\%.

%This strong outperformance across all ranks, as detailed in Table~\ref{tab:rank-n-iiit} and shown in Figure~\ref{fig:rank-n-plot}, highlights the effectiveness of our hybrid CNN-Transformer architecture.

\subsubsection{Results on Experiment \#2} 
The evaluation on the LFIW dataset, shown in Figure~\ref{fig:rank-n-plot-lfiw} and Table~\ref{tab:rank-n-lfiw}, highlights the challenges of fingerprint matching under real-world conditions. At Rank-1, our model achieves 2.86\% accuracy—competitive, though lower than NBIS (5.52\%). However, its strength becomes evident at higher ranks. As seen in Figure~\ref{fig:rank-n-plot-lfiw}, our model surpasses NBIS at Rank-4 and MCC at Rank-9. By Rank-10, it achieves 18.48\% accuracy, outperforming MCC’s 16.56\%. These results demonstrate the robustness of our hybrid architecture in handling cross-sensor and cross-surface variability. Notably, this performance is achieved despite training solely on the IIITD dataset, underscoring the model’s strong generalization capabilities.
\subsection{Ablation Study}

To evaluate the contribution of each part in the proposed LatentPrintFormer architecture, we conducted an ablation study with three setups: 1) CNN-only, 2) CNN with spatial attention, and 3) the full hybrid model with CNN, spatial attention, and Transformer. As shown in Table~\ref{tab:ablation-study}, the CNN-only model achieves a Rank-10 accuracy of 10.94\%. This result shows the basic effectiveness of local feature extraction. Adding spatial attention improves performance to 16.95\%. This change shows its ability to improve the important ridge areas while reducing background noise. The complete architecture, which includes the Transformer for global context modeling, raises accuracy significantly to 30.54\%. This trend emphasizes the combined benefits of spatial attention and Transformer-based long-range feature encoding for better latent fingerprint identification.

%%%%%%%%%%%%%%%%%%%%%%%%%%%%%%%%%%%%%%%%%%
\section{Conclusion}
\label{sec:conc}
This study presented a hybrid CNN and Transformer architecture designed to improve latent fingerprint identification. Experiments on the IIITD-Latent Fingerprint dataset demonstrated that the proposed model consistently outperformed established methods across all ranks.
To assess generalization, we evaluated the model on the LFIW dataset, which reflects real-world forensic conditions. Although all systems showed reduced performance, our model showed stronger improvement in higher ranks, indicating its effectiveness in narrowing down candidate lists, an important trait in forensic applications.
The overall results, particularly on LFIW, highlight that latent fingerprint identification remains a challenging problem. Future work should focus on enhancing robustness to sensor and environmental variations through improved enhancement techniques, unsupervised learning, and multi-modal approaches.
\section*{Acknowledgements}
This work was partially supported by Science Academies' Summer Research Fellowship Programme (SRFP) 2025 and Karnataka Power Transmission Corporation Limited (KPTCL). 
\bibliographystyle{splncs04}
 \bibliography{mybibliography}
\end{document}